\documentclass[sigconf]{acmart}

\usepackage{algorithmic}
\usepackage{textcomp}
\usepackage{xcolor}
\usepackage{balance}
\usepackage{hyperref}
\usepackage{stfloats}

\usepackage{booktabs}
\usepackage{graphicx}
\renewenvironment{quote}
  {\list{}{\rightmargin=0.4cm \leftmargin=0.4cm}%
   \item\relax}
  {\endlist}

\AtBeginDocument{%
  \providecommand\BibTeX{{%
    \normalfont B\kern-0.5em{\scshape i\kern-0.25em b}\kern-0.8em\TeX}}}

\copyrightyear{2022}
\acmYear{2022}
\setcopyright{acmlicensed}\acmConference[DIS '22]{Designing Interactive
Systems Conference}{June 13--17, 2022}{Virtual Event, Australia}
\acmBooktitle{Designing Interactive Systems Conference (DIS '22), June 13--17,
2022, Virtual Event, Australia}
\acmPrice{15.00}
\acmDOI{10.1145/3532106.3533536}
\acmISBN{978-1-4503-9358-4/22/06}

\acmSubmissionID{9970}

\begin{document}

\title{Designing for Caregiving: Integrating Robotic Assistance in Senior Living Communities}

\author{Laura Stegner}
\affiliation{%
  \institution{University of Wisconsin--Madison}
   \city{Madison}
   \state{Wisconsin}
   \country{USA}
  }
\email{stegner@wisc.edu}

\author{Bilge Mutlu}
\affiliation{%
  \institution{University of Wisconsin--Madison}
   \city{Madison}
   \state{Wisconsin}
   \country{USA}
  }
\email{bilge@cs.wisc.edu}


\begin{abstract}

Robots hold significant promise to assist with providing care to an aging population and to help overcome increasing caregiver demands. Although a large body of research has explored robotic assistance for individuals with disabilities and age-related challenges, this past work focuses primarily on building robotic capabilities for assistance and has not yet fully considered how these capabilities could be used by professional caregivers. To better understand the workflows and practices of \textit{caregivers} who support aging populations and to determine how robotic assistance can be integrated into their work, we conducted a field study using ethnographic and co-design methods in a senior living community. From our results, we created a set of design opportunities for robotic assistance, which we organized into three different parts: supporting caregiver workflows, adapting to resident abilities, and providing feedback to all stakeholders of the interaction.
   
\end{abstract}

\ccsdesc[500]{Human-centered computing~User studies}
\ccsdesc[500]{Human-centered computing~Field studies}
\ccsdesc[500]{Computer systems organization~Robotics}

\keywords{human-robot interaction, care robot, caregiving, field study, design recommendations}

\begin{teaserfigure}
  \includegraphics[width=\textwidth]{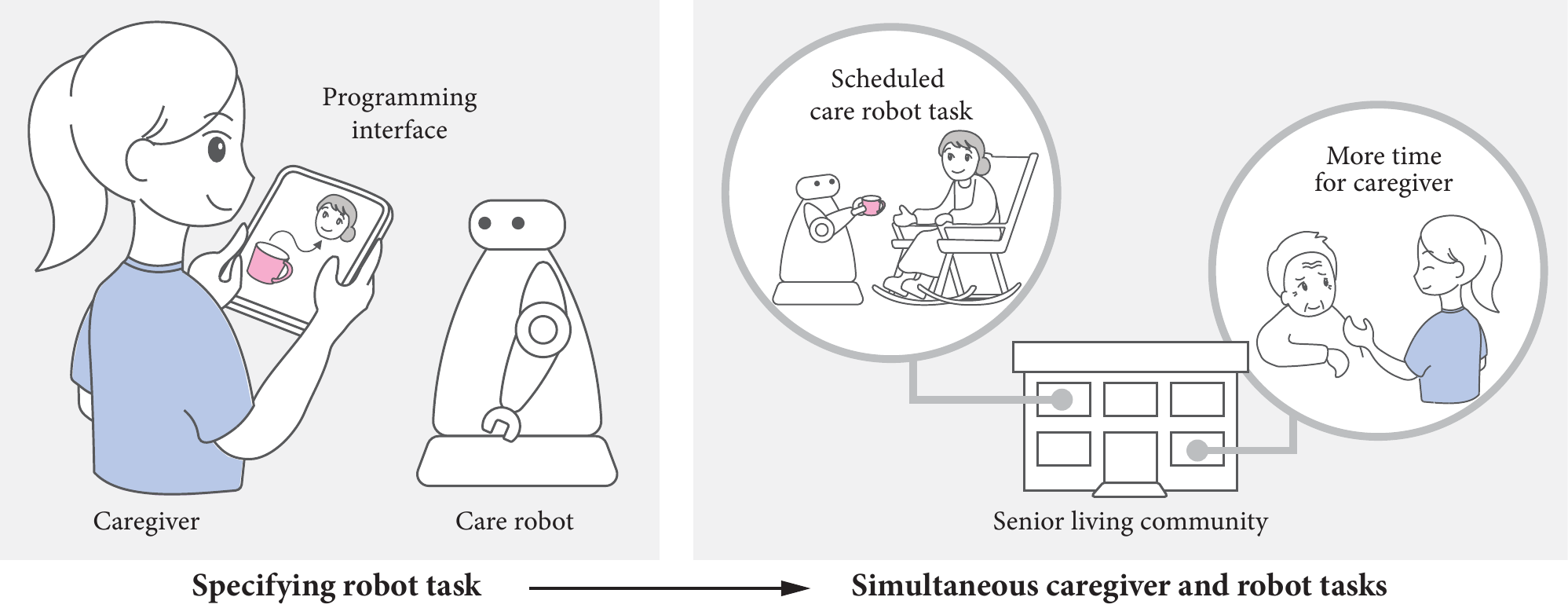}
  \caption{We conducted a field study to find opportunities for robots to support caregivers in assisted and independent living settings. The figure illustrates a potential scenario where a caregiver can specify routine tasks for the robot to perform. The caregiver can then engage in more meaningful interactions with residents while the robot completes more mundane tasks.}
  \Description{On the left, a caregiver is holding a tablet and standing facing a robot. The label says "programming interface." On the right, two scenes in a senior living community are shown. The first is labeled "scheduled care robot task" and shows a robot delivering a coffee cup to an elderly woman sitting in a chair. The second is labeled "More time for caregiver" and shows a caregiver smiling at an elderly man.}
  \label{fig:teaser}
\end{teaserfigure}

\maketitle

\section{Introduction}
The population is aging globally and across nations, with the proportion of adults in the U.S. aged 65 and older projected to grow from 13.1\% in 2010 to 21.4\% in 2050 \cite{pew2014aging}. This growth will cause old-age dependency, i.e., the number of people older than 64 per 100 people of working age, to nearly double from 19 to 36 \cite{pew2014aging}. Supporting independent living for this population will require a substantial increase in care services, although the caregiver workforce is not growing at a rate that can meet this need. A shortage of 355 thousand caregivers is expected by year 2040 in the U.S. \cite{famakinwa2021report}. Technology presents many opportunities to help close this gap, although the existing technological landscape focuses on managing, scheduling, and monitoring care workers rather than performing essential care tasks \cite{okeke2019technology}. Autonomous robots hold significant promise in addressing this gap through capabilities including mobility, manipulation, and learning. However, these capabilities must be designed carefully for robots to perform tasks that they are best suited to perform, to work in harmony with caregivers, and to be accepted by older adults.
Our work aims to help build an understanding of caregivers' work to inform the design of care robots to support caregivers in scenarios such as the use case highlighted in Figure \ref{fig:teaser}.

Caregivers support individuals in sustaining and enjoying life. In order to live independently, individuals must be self-sufficient with both \textit{Activities of Daily Living (ADLs)}, which include basic personal tasks such as bathing, dressing, using the toilet, eating, ambulating, or transferring to or from a bed or chair, as well as \textit{Instrumental Activities of Daily Living (IADLs)}, which include more complex planning and thinking such as housework, taking medication, preparing meals, shopping, and using communication devices \cite{spector1998combining}. As people age, most will eventually require some form of assistance \cite{thomas2015long}. Depending on the level of care required, aging individuals may be moved into a \textit{senior living community}, which includes facilities that support \textit{Independent Living (IL)} or \textit{Assisted Living (AL)}. IL facilities provide ``light'' assistance with IADLs and possibly one or two ADLs such as dressing or bathing, but the residents are almost completely independent and do not need assistance with ADLs such as transferring, ambulating, or using the toilet \cite{perkins2004building}. For example, a resident may require assistance with managing medication and getting dressed in the morning but can otherwise perform tasks necessary to be independent. In contrast, AL offers support at all hours to assist with a range of ADLs and IADLs \cite{zimmerman2007definition}. Residents in AL can expect assistance with a range of activities from getting out of bed in the morning to meal preparation and cleanup as well as access to help with unscheduled needs such as using the toilet \cite{kane1993assisted}. 

A wealth of research in the last two decades has explored how autonomous \cite{law2021case,schaeffer1999careobot} and teleoperated \cite{chen2013robots,michaud2007telepresence} robots can directly deliver care to individuals in need. This body of literature has explored the specific needs of people with disabilities or age-related challenges, such as difficulty bathing due to limited mobility \cite{king2010towards}, and has developed robotic solutions that can address these needs, including assisting individuals with ADLs and IADLs \cite{chen2013robots,luperto2019towards}. The development of such capabilities is critical to realize the vision of care robots, but how these capabilities will be utilized by caregivers and how such robots can be integrated into day-to-day care routines and workflows remains relatively under-explored.

To determine how a robot could assist caregivers with their work and to uncover opportunities for robot design, we conducted a field study using ethnographic and co-design methods with caregivers in a senior living community. First, we observed caregivers during their shift with fly-on-the-wall observations to gain contextual insight into their tasks and workflows. Second, we conducted interviews with those caregivers to supplement the observations. The interviews also included co-design activities toward developing an understanding of the caregivers' perspectives on how a care robot could support their work. We report on our findings from an analysis of the resulting data and discuss their implications for the integration of care robots into care routines and workflows.

Our work makes the following contributions:
\begin{itemize}
    \item A better understanding of how caregivers in AL and IL settings work, including characterizations of day-to-day routines and workflows, through the lens of robotic assistance;
    \item A set of design implications for robotic technologies in senior living communities, including supporting caregiver work-flows, adapting to resident abilities, and providing feedback to all stakeholders of the interaction.
\end{itemize}

\section{Related Work}\label{sec:related}
\subsection{Tasks and needs of caregivers}
Prior work in gerontology has developed a strong understanding of how caregivers should provide care to residents in senior living facilities.
Training manuals \cite{somers2008caregivers,garrod2020advanced} provide detailed guidelines on assisting individuals with ADLs and IADLs, as well as general interaction considerations such as communicating with someone with cognitive decline and preventing falls.
More specialized studies have analyzed specific facets of caregiving, such as the need for personalization of care \cite{miller2021implementation}, importance of caregiver training \cite{falk2017need}, balancing physical setting with social and organizational context \cite{zimmerman2001assisted}, creating a welcoming environment \cite{johnston2019welcoming}, planning effective events \cite{fu2015insights}, and creating positive family-staff relationships \cite{bauer2011improving}.
Additional work been done to develop ethics frameworks for resident-focused issues in everyday settings \cite{kemp2021ethics,powers2005everyday}.

While caregiving practices have been widely studied, the industry suffers from burnout \cite{chan2021caregiving}.
In an effort to better assist caregivers in their day-to-day jobs, \textit{Ambient Assisted Living} (AAL) systems are increasingly used to help monitor residents in care facilities or at home \cite{rashidi2013survey} using a combination of smart home sensors \cite{ghayvat2018smart} and wearable technologies \cite{marques2019ambient}.
However, \citet{offermann2018they} found that professional caregivers, particularly of disabled people, were critical of AAL systems and their designs, particularly regarding the potential for continuous monitoring equipment such as cameras and microphones to violate privacy and human dignity. 
Several works \cite{aced2015supporting,zulas2012caregiver} have shown success with including caregivers in the design of these AAL systems, pointing to the need to closely consider caregiver needs and perspectives when designing these kinds of technologies.

The experience and burden of informal caregivers who care for family or friends has also been widely studied \cite{chen2013caring,montgomery1985caregiving,grunfeld2004family}. Their burden is often considered in two classes: \textit{objective}, meaning the tasks the caregiver must perform for the care recipient, and \textit{subjective}, meaning the emotional toll that comes with providing the care \cite{jones1996association}. \citet{montgomery1985caregiving} found that while objective burden can be eased through interventions that free the caregiver's time, subjective burden is often linked to factors such as age and income that are not easy to change. Systems such as Ambient aNnotation System (ANS) \cite{quintana2013augmented} and CareNet Display \cite{consolvo2004carenet} have been developed to ease the objective burden of informal caregivers at home. While formal and informal caregivers face different challenges with their work, they share a similar objective burden, such as through the care tasks performed and the need to monitor care recipients.

\subsection{Existing care robots}
Researchers have developed a number of care robots to address the needs and expectations of older and clinical populations. Systems such as Care-o-Bot \cite{schaeffer1999careobot}, PR2 \cite{chen2013robots}, and Hobbit \cite{fischinger2016hobbit} were designed to provide general assistance to care recipients, including manipulators that allow for interaction with the environment. Other work has focused on mobile robots for monitoring and promoting safety and well-being by integrating robots with smart environments and sensors \cite{noury2005ailisa,badii2009companionable,gross2015robot,nani2010mobiserv}. While these robots are mainly autonomous \cite{pollack2002pearl,graf2004careobot,nani2010mobiserv,dario1999movaid,schaeffer1999careobot}, some systems are focused on teleoperation and telepresence for a caregiver to communicate with a resident remotely \cite{chen2013robots,michaud2007telepresence,luperto2019towards}.
Commercial application of care robots has also gained support recently, with companies such as Pal Robotics,\footnote{Pal Robotics: \url{https://pal-robotics.com/}} F\&P Robotics,\footnote{F\&P Robotics: \url{https://www.fp-robotics.com/}} Diligent,\footnote{Diligent: \url{https://www.diligentrobots.com/}} and Labrador\footnote{Labrador: \url{https://labradorsystems.com/}} marketing robots gear toward general home assistance applications.

In addition to developing technical capabilities, studies of these systems have assessed their effectiveness in care task performance and care recipient perceptions. For example, \citet{schroeter2013realization} deployed the Hector robot in a smart home environment to assist older adults with cognitive impairments for a period of time. While care recipients found the robot useful and enjoyable, family members expressed the desire to set up and control the robot \cite{schroeter2013realization}. 
This study highlighted the importance of considering caregivers in addition to care recipients in care robot design. 

This impressive array of systems shows the feasibility of robotic assistance in care settings and helps outline the design space for care robots. 
They represent significant technological advancements that address long-term care needs, with particular focus on providing effective assistance and creating positive experiences for the resident. However, results from field study deployments show that current caregiver needs are not sufficiently considered in terms of personalized care practices and integration in existing workflows.

\subsection{Designing care robots with stakeholders}
A sizable number of studies have aimed to develop design requirements for autonomous and teleoperated robots for care settings. These studies use methods such as participatory design sessions \cite{eftring2016designing,sabanovic2015robot,winkle2018social}, ethnographies \cite{forlizzi2004assistive,pirhonen2020could}, interviews \cite{beer2012domesticated, law2019developing}, and focus groups \cite{badii2009companionable,michaud2010exploratory} to understand the needs of older adults living independently and to support autonomy among older adults.
Other studies explored how robots can provide assistance in retirement communities and attitudes toward robots through questionnaires and interviews with residents, family, and staff \cite{broadbent2009retirement,broadbent2012attitudes}.
Additional studies have looked at how care robots can be used to support informal caregivers, such as family, as they manage care needs in addition to their own lives \cite{moharana2019robots,berry2017values}.
All of these studies show how different design approaches with various stakeholders in care robots can create a more complete understanding of care robots.

While much is known about caregiver workflows, less is known about how we can integrate care robots into their workflow successfully. Several studies have begun to examine how care robots can be integrated into care environments and workflows. 
For example, \citet{bardaro2021robots} discussed limited adoption of care robots despite technical developments, recommending a co-design approach to identify specific needs that robots can address. 
Similarly, \citet{alaiad2014determinants} identified factors that affected ``usage intent'' and found that the caregivers and care recipients had different preferences regarding what tasks the robot should perform.
Finally, \citet{hornecker2020interactive} conducted an ethnographic study of practices regarding a robotic lifting device in gerontological care to identify ways of better integrating  robots into these care environments, recommending the consideration of \textit{triadic} interactions involving resident, caregiver, and robot systems instead of \textit{dyadic} interactions involving care recipients and robots. We seek to build on this work by considering more versatile robots and consider the triadic nature of these interactions in our design implications. Prior work illustrates the present need to consider how care robots fit into current caregiver workflows, rather than considering them as independent agents.


\section{Method}
To identify how caregivers might benefit from care robots, we conducted a field study using ethnographic and co-design methods at a senior living facility that offered both independent and assisted living services. We intermittently conducted onsite observations and interviews with caregivers from both care settings during August--September 2021. All study methods were reviewed and approved by an institutional review board (IRB).
This study took place during the COVID-19 pandemic, which caused high rates of turnover and frequent pauses to the study due to outbreaks in the facility, thereby negatively impacting the number of participants we were able to work with. Researchers adhered to all regulations of the facility.

\subsection{Research Context}
We collaborated with a senior living facility, which we refer to as ``facility'' to protect participant confidentiality.
The facility is suburban, private, not-for-profit and located in the Midwestern United States. It includes 85 Independent Living (IL) apartments and 60 Assisted Living (AL) apartments. The IL section is staffed by two caregivers during the day, one during the evening, and one on-call during the night. The AL section, has caregivers available at all hours with at least one caregiver per ten residents, which is slightly higher than typical caregiver-to-resident ratios.

\subsection{Data Collection}

\subsubsection{Participants}
In total, seven caregivers, aged 29--64 ($M=50.0$, $SD=12.9$; all female), participated in the study. 
This skew in participant gender is expected since the majority of healthcare workers (79--89\%) are women \cite{argentum2018senior}. 
Participants' caregiving experience varied between 1 month to 26 years ($M=11.8$ years, $SD=9.96$ years). Two participants opted out of sharing demographic and experience data. Of the seven participants, three worked in AL only, three worked in IL only, and one worked in both. Table \ref{tab:participation} shows caregiver participation, which included a total of 13 sessions.
Participants received a flat fee of \$20 USD to be observed and \$40 USD/hour to participate in interviews as compensation.

				


\begin{table}[!t]
\caption{Caregiver participation in study activities.}
\label{tab:participation}
\centering\small\renewcommand{\arraystretch}{1.1}
\begin{tabular}{ ll } 
    \toprule
    \textbf{Study Session} & \textbf{Participant} \\ 
    \midrule
    AL Observation (day) & B1 \\
    AL Observation (partial pm) & AL1 \\
    AL Observation (partial pm) & AL2 \\
    IL Observation (pm) & B1 \\
    IL Observation (pm) & IL1 \\
    IL Observation (day) & IL2 \\
    IL Observation (day) & IL2 \\
    IL Observation (partial day) & IL1 \\
    Interviews & AL2, AL3, IL1, IL2, IL3 \\
    \bottomrule
\end{tabular}
\end{table}

\subsubsection{Observations}
The goal of the observations was to understand caregivers' main tasks and workflows. Observations provide valuable information about the natural context and workflow structures, and they reveal ``tacit knowledge'' \cite{polanyi2009tacit} that is relevant to human-robot interaction design. To the extent that it was possible due to privacy concerns of residents, we conducted fly-on-the-wall observations in order to minimally affect the observed workflows. Because the nature of the caregiver's work involved entering the private rooms of residents, we obtained permission to observe the care interaction from each resident. If a resident declined, the researcher waited outside of the room while the caregiver assisted that resident. During some observations, the researcher inadvertently participated in care activities, for example, by holding materials. Observations lasted for either half or all of the caregivers' normal shifts. To protect the privacy of residents, we only took field notes. 

\subsubsection{Interviews}
After the observations were completed, we interviewed caregivers during separate study sessions with the goal of understanding the caregivers' view of their work and its challenges. Additionally, we gathered caregivers' ideas about how a robot might assist with their work. Interviews were semi-structured, including:
\begin{enumerate}
 \item demographic questions about their work experience;
 \item an overview of their typical day;
 \item challenges they associate with their work; 
 \item how an untrained human assistant can help with their work;
 \item their general attitude of and expectations for robots;
 \item how they imagine a robot can help with their work; and
 \item challenges they foresee with a robot in the care facility. 
\end{enumerate}
With question six, we provided participants a paper and multicolored pens and asked them to sketch what they would want a robot that helps them to look like. The sketch served as a prompt for us to ask questions regarding the robot's features, abilities, and duties. 
After the sketches were discussed, we then presented a set of images of robots to the caregiver, including the Stretch \cite{kemp2021design}, PR2 \cite{chen2013robots}, Talos \cite{stasse2017talos}, and Lio \cite{mivseikis2020lio}. We chose these particular robot images to inspire more creativity among the caregivers. When selecting the visual prompts, we selected robots with different form factors but roughly similar abilities: manipulation, mobility, vision, and hearing. 
Each robot image was presented individually, and the caregiver was asked to describe what the robot should do to help them. Our focus in the interviews was to understand what care robots need to do to be useful to the caregivers, so we did not discourage unrealistic beliefs about robot abilities. 
Instead, the researcher used their human-robot interaction knowledge to probe the caregiver about their design choices. The caregivers also asked clarification questions to the researcher to better understand the robot abilities. 
Each interview lasted 30--60 minutes and was conducted in a private, quiet room at the facility. We recorded audio and video data that we transcribed for analysis.

\subsection{Data Analysis}
All data, including field notes and interview transcriptions, were standardized and unitized in text form. 
We analyzed the data using applied thematic analysis following the guidelines by  \citet{boyatzis1998transforming} and \citet{guest2011applied}.
From the data collection process, we were already familiar with the data prior to beginning analysis. We first identified preliminary themes by reading the data and identifying points of potential significance relating to the research objective. 
Then, we assigned codes to significant references and events during an iterative coding process. The codebook was modified ``as new information and new insights are gained'' \cite{guest2011applied}. 

After the codebook was finalized, we trained a secondary coder to assess inter-rater reliability (IRR). After training, the secondary coder used the code book to assign codes to 10\% of the data. Reliability analysis indicated ``almost perfect'' reliability according to interpretation guidelines provided by \citet{landis1977measurement} (Cohen's Kappa, $\kappa=0.89$). We resolved disagreements through discussion.
Once the coding and IRR analysis was complete, we revised the preliminary themes based on support from the codes and data.

To gain a better understanding of the caregiver workflows, we additionally analyzed our field notes using principles from social science framing \cite{lofland1971analyzing}. 
Social science framing is not a strict procedure, but instead provides considerations for how to organize qualitative data into social and temporal relationships. We used these considerations to identify significant events that shape the flow of the caregiver's shift, such as identifying regular \textit{practices}; brief, unexpected \textit{encounters}; and longer, unplanned \textit{episodes}. These results are presented in the form of a reconstituted timeline of events. 
\begin{table}[!t]
\caption{A summary of the themes from our analysis.}
\label{tab:summary}
\centering\small\renewcommand{\arraystretch}{1.1}
\begin{tabular}{ p{0.95\linewidth} } 
    \toprule
    \textbf{Summary of Findings} \\ 
    \midrule
    \textit{Factors that Shape Caregiving} \\ 
    \textbf{Theme 1: Caregiver workflows} \\ AL and IL caregivers have scheduled tasks, but AL has an unpredictable workflow with interruptions. Time management is a common challenge. \\
    \textbf{Theme 2: Resident needs and preferences} \\ Day-to-day interactions with residents differ based on each resident's abilities, routines, and preferences, which caregivers learn over time. \\
    \textbf{Theme 3: Communication} \\ Caregivers actively maintain transparency with residents and communicate with each other by documenting care thoroughly. \\
    \midrule
    \textit{Desired Role of the Robot} \\
    \textbf{Theme 4: Providing physical support} \\ Caregivers envision a mobile humanoid robot that performs care tasks for residents and detects hazards, such as damp materials or smoke. \\
    \textbf{Theme 5: Providing mental and emotional support} \\ Companionship and comfort are critical to resident care. Robots should monitor residents' mental states but not provide this social support. \\
    \textbf{Theme 6: Expectations of interaction modality} \\ Caregivers want robots to handle a mix of scheduled tasks and interruptions. Robots should be overseen by caregivers for resident safety. \\
    \bottomrule
\end{tabular}
\end{table}

\section{Results}
In our analysis, six major themes emerged about how caregivers work and what they desire from a care robot, which we group into two high-level categories for clarity. The themes are summarized in Table \ref{tab:summary}. For each theme, we first provide a high-level description, and then present supporting quotes from the interviews and observations. Both quotes and observations are attributed using participant ID. We made minimal edits and added annotations to the quotes to improve clarity while retaining the meaning.
Study data is available via OSF.\footnote{Study data and materials are available through the following \textit{OSF} repository: \url{https://osf.io/mfkr5/?view_only=4ce32ce172e34c5eab618f654e79c4ed}}

\subsection{Factors that Shape Caregiving}
Our study revealed a number of factors that impacted how caregivers work and what considerations they have while assisting residents. These factors come from a combination of caregiver comments in the interviews and our observations during shifts. While AL and IL care practices share many similarities, we highlight the key differences we observed between them for each factor.

\subsubsection{Theme 1: Caregiver workflows}
\begin{figure}[!t]
    \centering
    \includegraphics[width=\columnwidth]{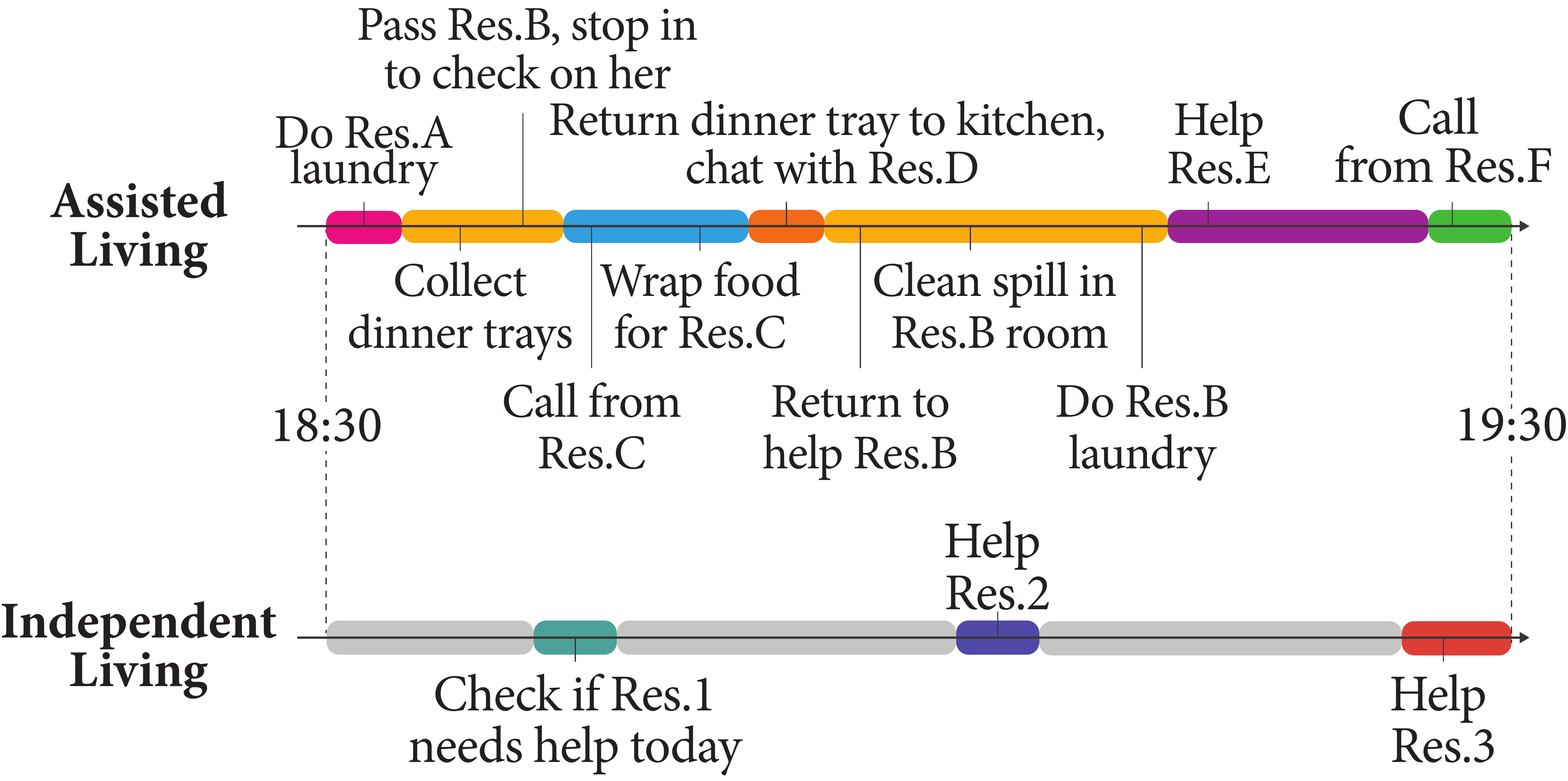}
    \caption{Assisted living and independent living caregivers have drastically different workflows. In AL, caregivers are constantly switching between residents in an on-demand style. IL caregivers tend to have a more fixed schedule. The colors above indicate when a caregiver is with a specific resident, and grey denotes caregiver downtime in between tasks.}
    \Description{Two timelines stacked vertically, labeled Assisted Living and Independent Living. Both timelines start at 18:30 and end at 19:30. The Assisted Living timeline contains the following colored labels from left to right: pink (Do Res A laundry), yellow (collect dinner trays; pass res B, stop in to check on her), blue (call from res C, wrap food for res C), orange (return dinner tray to kitchen, chat with res D), yellow (return to help res B, clean spill in res b room, do res b laundry), purple (help res e), and green (call from res f). The Independent living timeline contains the following colored labels from left to right: grey (no label), teal (check if res 1 needs help today), grey (no label), navy (help res 2), grey (no label), and red (help res 3).}
    \label{fig:comparison}
\end{figure}

Our analysis shows that in terms of task predictability, AL and IL workflows differ greatly. Caregivers in both settings have assigned tasks and encounter unexpected situations that need attention. However, specific day-to-day routines vary greatly from one setting to another. The AL setting has a more unpredictable workflow than IL, as visualized in Figure \ref{fig:comparison} by an exemplar workflow in AL versus IL. While we expected to see such differences in workflow based on previous work that aims to classify these settings \cite{kane1993assisted,zimmerman2007definition,perkins2004building}, our work presents the opportunity for a more detailed account because of how these workflows can impact robot design.

\begin{figure}[!b]
    \centering
    \includegraphics[width=\columnwidth]{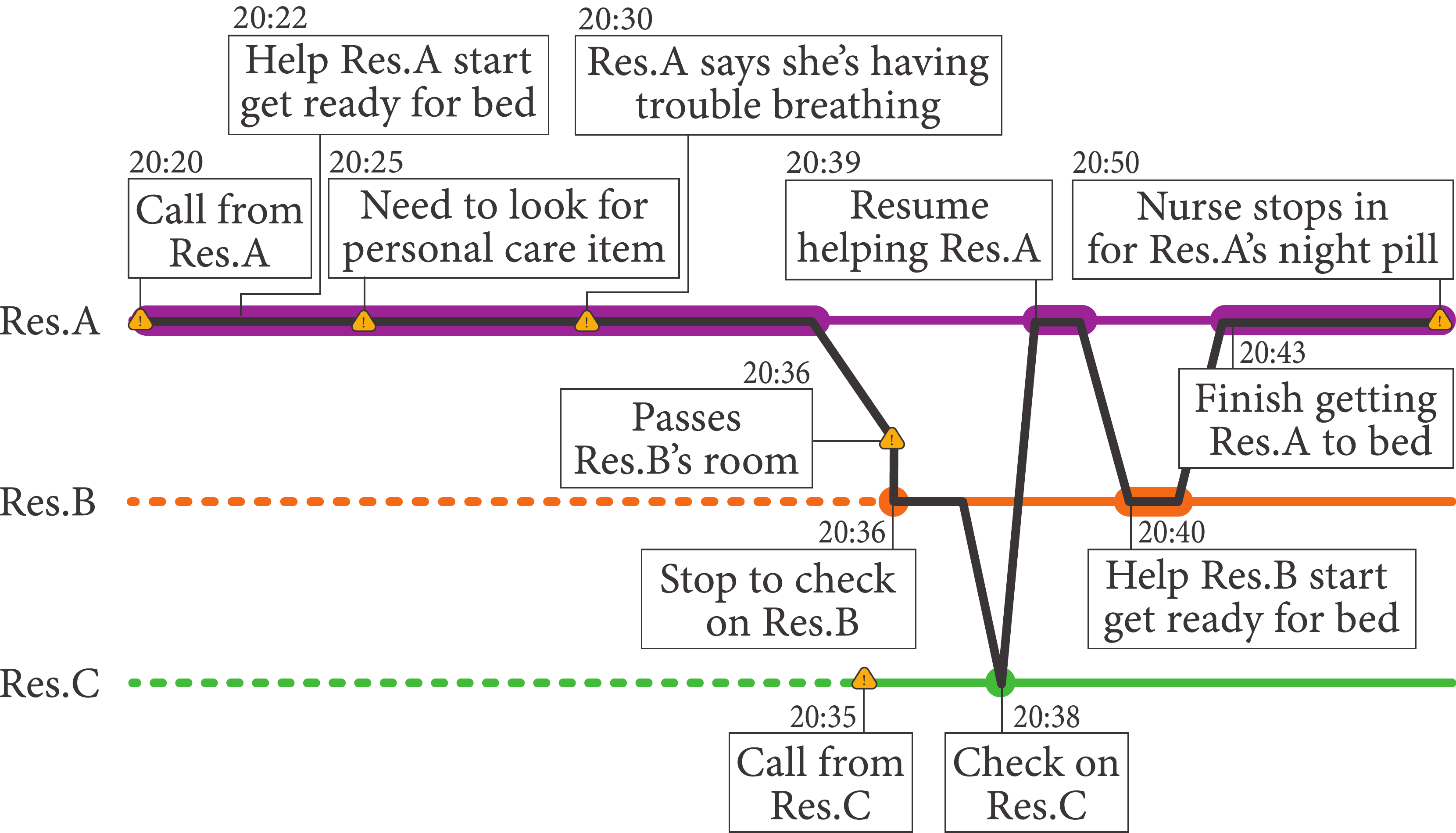}
    \caption{Caregivers in AL face significant interruptions in their work. One example is shown here, where the caregiver is helping Resident A get ready for bed. The black line shows the path the caregiver takes trying to help Resident A, but also assist Residents B and C. Residents can be left waiting for a caregiver to return because of their large workload.}
    \Description{Three timelines stacked vertically, labeled Res A, Res B, and Res C from top to bottom. A thick line follows Res A's line, then goes down to Res B, then to Res C, then back up to Res A, then back to Res B, and finally ending at Res A. Significant events are labeled on the timelines. The labeled events from left to right read "20:20 call from Res A;" "20:22 Help Res A start getting ready for bed;" "20:25 Need to look for personal care item;" "20:30 Res A says she's having trouble breathing;" "20:35 Call from Res C;" "20:36 Passes Res B's room;" "20:36 Stops to check on Res B;" "20:38 checks on Res C;" "20:39 resume helping Res A;" "20:40 Help res B start get ready for bed;" "20:43 Finish getting Res A to bed;" and "20:50 Nurse stops in for Res A's night pill."}
    \label{fig:interruptions}
\end{figure}

\paragraph{Assisted Living. }
In the AL setting, caregivers encounter numerous interruptions, which require them to tend to multiple competing requests. These interruptions arise when the caregiver either observes something unexpected that they need to investigate, such as a potential hazard, or when they are paged by a resident in need of assistance. We observed that residents were often left waiting on assistance from caregivers, and that the caregivers often had to leave them mid-task due to interruptions, as shown by the timeline of field note observations in Figure \ref{fig:interruptions}. 

\paragraph{Independent Living. }
The IL setting, on the other hand, follows a much more structured, predictable workflow. The caregivers have scheduled times to assist residents. They ``\textit{go at different times through the day}'' (IL3) to assist the resident with ``\textit{scheduled}'' (IL1,IL2) daily tasks such as taking medication and getting dressed. While emergencies can demand their attention away from their scheduled work, such interruptions are ``\textit{rare}'' (AL3). 

\paragraph{Commonalities. }
Despite the workflow differences, all of the caregivers mentioned time management as a challenge, particularly when multiple residents need help at once. One common thread was the idea of prioritizing tasks based on urgency. The caregivers ``\textit{always prioritize the bigger things}'' (AL2) such as toileting, rather than smaller tasks such as delivering ice water. They also need to respond quickly to emergency calls because they ``\textit{don't know what's going on}'' (IL3); the resident could be ``\textit{bleeding on the floor}'' (IL3) or ``\textit{really upset}'' (IL1). This prioritization can cause some residents with non-urgent requests to be left waiting and potentially unhappy because ``\textit{small things are super important to them}'' (AL2).

\subsubsection{Theme 2: Resident Needs and Preferences}
While at a high level, the caregivers of AL and IL have the same qualifications and training, their day-to-day interactions with the residents vary greatly between these two facilities. Prior work has emphasized the need for personalized eldercare \cite{miller2021implementation}, which we saw reflected in the way that caregivers addressed individual resident needs and preferences. Here, we report how this aspect needs to be approached when designing care interactions between residents and robots.

\paragraph{Abilities. }
Residents in AL required more assistance, such as toileting, transferring, and ambulating, whereas residents in IL are ``\textit{more independent, and they want to stay that way}'' (IL3). They only required light physical assistance with tasks such as taking medication, bathing, or changing clothes. The range of physical and mental abilities observed among residents in AL and IL matched prior work in this space \cite{perkins2004building, zimmerman2007definition}. 

In both AL and IL, residents can have physical deficits, such as hearing loss or low vision. The field researcher observed instances where the caregiver had to adjust her behavior to accommodate a hard-of-hearing resident. For example, it was noted from observing AL1 that ``\textit{Resident is very hard of hearing, so AL1 is talking loudly, directly in her ear.}'' This kind of behavior was observed from both AL and IL caregivers.
Further, residents might suffer from mental deficits, such as memory problems or confusion. AL3 describes how she customizes her care for residents who forget to use the call button to request assistance, saying ``\textit{There's a couple that just never ever use a call light, but you know you need to check on because they're compromised cognitively, and bizarre things happen, you just need to be very mindful of their well-being with their whereabouts and things like that.}''

\paragraph{Routines. }
In addition to individual resident needs, both AL and IL caregivers expressed the importance of a resident's individual routine. 
In AL, this knowledge of routine proved useful for the caregivers when planning their shifts and understanding normal resident behaviors. For example, AL2 describes how she uses knowledge of her residents' routines to plan for bedtime, saying `\textit{`they all go to bed around the same time, and if we've been with them for long enough, we know ... the order to put them in bed.}''
In IL, the caregivers consider routine from the perspective of timeliness being important to the residents. If the caregiver is late, the resident will worry or potentially be upset about having to wait. IL1 highlights that ``\textit{their days can be really really long, so they're on a schedule.}'' The residents are ``\textit{expecting}'' (IL3) the caregivers, and can be upset even if the caregiver is only ``\textit{five minutes late}" (IL1).

\paragraph{Preferences. }
Residents also have specific individual preferences, which caregivers learn over time and use to anticipate a resident's desires and to prevent them from repeating requests. 
However, these resident preferences are not written down anywhere, meaning that each caregiver has to learn them over time. AL2 explains this by saying, ``\textit{The kind of blankets they like to put on, and the order of it and ... how they get situated in bed ... they don't put that in ... their medical record.}'' IL2 echoed a similar sentiment, saying ``\textit{[The residents] have a routine, and sometimes they don't even know what the routine is until you start working together, then they figure out, `Well I like it this way,' ... so you want to make them happy. You want to make them comfortable.}''

\subsubsection{Theme 3: Communication}
Caregivers use a diverse set of communication strategies when interacting with residents and with each other. Considering these communication methods is key to allowing care robots to fit into this social environment.

\paragraph{Communication With Residents.}
The caregivers communicate directly with residents by being transparent with residents about the caregivers' actions and intentions and by listening to residents dictate to the caregivers what they want or need.
When providing care, the caregiver takes additional steps to include the resident in the process. The caregiver asks permission to do tasks and informs the resident of what is being done. This transparency was observed frequently in field notes in both AL and IL. For example, when observing B1 in AL, we noted ``\textit{B1 says `I'm gonna straighten you out, okay?' and the resident replies `Atta girl, use your muscles.'}\thinspace'' and noted that ``\textit{IL2 takes the resident's temperature, [then] says it to the resident.}''
Residents also took initiative to communicate with the caregivers. Particularly in AL, the residents were not shy about instructing the caregiver about how to perform certain tasks. 
The field notes record an instance of these instructions, noting that ``\textit{Resident gives AL2 a to-do list before bed: leave night light on, clean catheter bag, close closet door.}''
While residents had call buttons that would summon a caregiver, they typically only used them for urgent or emergent situations and saved small requests to be communicated once the caregiver was present for another reason.

\paragraph{Communication Between Caregivers.}
Caregivers communicate with each other formally through an electronic charting system that allows them to track what assistance they provide to residents, as well as notes about their general health and well-being. The caregivers ``\textit{write down the ... services [they] provide, [and] anything that was out of the ordinary}'' (AL3). IL3 explains the importance of these notes, saying ``\textit{because the [previous caregiver] leave[s] before I get here, so it's how we communicate. They leave me a note.}''
All caregivers were observed charting during their shifts. AL caregivers typically did all of their charting at the end of the shift, whereas IL caregivers typically charted after each resident.

AL and IL caregivers had different styles of interpersonal communication.
AL caregivers had much more interaction with other caregivers they would see in passing. They were observed to stop to chat briefly about a resident or general information about the facility. 
Since IL caregivers are not working on such a large team, they do not have these brief interactions.
However, both AL and IL caregivers emphasized ``\textit{how good it is to work as a team}'' (AL2). IL2 describes that she can ``\textit{call for help}'' from other staff members to help her out if she is behind.

\subsection{Desired Role of the Robot}
Caregivers provided various insights into how a care robot could assist with their work. While we noted differences between the roles of AL and IL caregivers, we did not find noteworthy differences in how they envisioned a robot assistant. They gave feedback surrounding the physical and emotional capabilities of the robot, as well as their expectations of interaction modality. All interviewees expressed that they were open to the idea of robots assisting them, but none had experience with robots outside of seeing them in entertainment or other media. They each voiced a desire to ``one-on-one meet'' (IL2) a robot and ``see where their limits lie'' (AL3).

\begin{figure}[!t]
    \centering
    \includegraphics[width=\columnwidth]{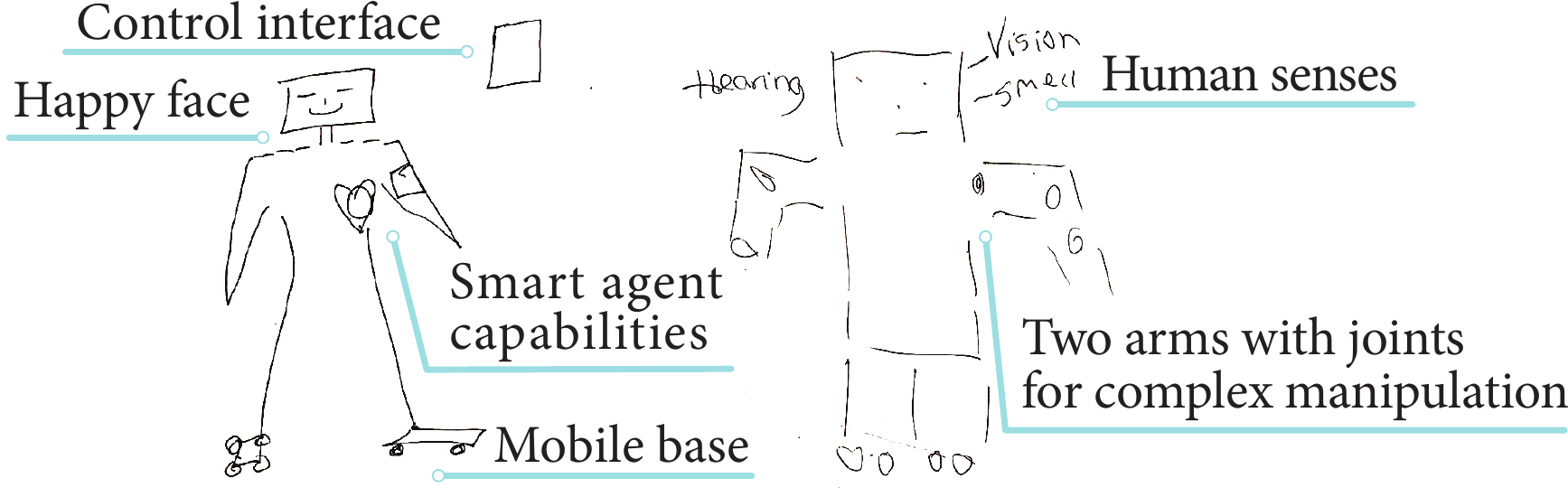}
    \caption{During the interviews, caregivers created sketches of how they envision a robot that could help with their work. All sketches portray highly capable robots with two arms and a mobile base. 
    Caregivers highlighted the need for standard sensing capabilities such as vision and hearing, as well as suggesting more novel capabilities such as taste, smell, and touch. Other features included emotional intelligence, incorporation of a smart agent, and a control interface.}
    \Description{Two hand-drawn sketches of humanoid robots. Each sketch is labeled by significant feature. On the left, "control interface" points to a square beside the robot, "happy face" points to the smile on the robot's face, "smart agent capabilities" points to a heart drawn on the robot's chest, and "mobile base" points to wheels drawn on the feet of the robot. On the right, "human senses" point to labels on the robot's face indicating hearing, vision, and smell; and "two arms with joints for complex manipulation" points to the robot's arms, which include joints for movement.}
    \label{fig:robot_sketches}
\end{figure}

\subsubsection{Theme 4: Providing Physical Support}
The caregivers expressed desire for highly capable robots to perform complex physical tasks. We did not constrain their discussion to existing robot capabilities, opting instead to explore caregiver expectations for future robots.

\paragraph{Physical tasks.}
As part of the interview, caregivers were asked to sketch their vision of a care robot that could assist with their work, as shown in Figure \ref{fig:robot_sketches}. While the caregivers created simple drawings, these drawings prompted in-depth discussions about the envisioned robot's appearance and capabilities.

All of the caregivers drew a humanoid robot with two arms, a smiling face, and a mobile base. While they indicated two arms would be more useful, when shown images of single-arm collaborative robots, the caregivers could still see some value and use for them. IL1 explains her preference, saying ``\textit{When I say that it could do what I could do, I have two hands and arms, you know. I think it could do more, and that's why I love the ones with the two hands better.}'' IL1 goes on to explain why she drew a smiling face, saying ``\textit{I would expect it to have a face, too, and a happy one, because I think we all need some happiness in our life. I don't think it should be too industrial at all.}''
The caregivers in AL also described a robot that had ``\textit{the capacity to lift}'' (AL3) residents, such as ``\textit{move a limb}'' (AL3) or ``\textit{lifting them up ... off the toilet}'' (AL2). While the IL caregivers did not mention lifting residents, IL3 stated the robot could ``\textit{move a table.}''
Finally, the caregivers described the desire for a ``\textit{waterproof}'' (AL2) robot, enabling the robot to help with tasks such as washing dishes or bathing a resident.  

Other considerations that emerged relate to the environment in which the care robot would operate. For example, the field notes report that some residents require oxygen, so they have delicate machinery set up and long oxygen tubes running through their living space. Damaging any of the tubes can cause a health hazard for the resident. Additionally, the caregivers expressed wanting a robot to ``\textit{clean up spills}'' (AL2) and that ``\textit{infection control would be huge ... especially in these days with COVID}'' (AL3).
Finally, due to the home-like environment of the residents' rooms, caregivers state that the robot needs to be able to get ``\textit{in smaller spaces}'' (AL2). Furthermore, residents often move slowly, whether it is walking or moving in a wheelchair. IL3 expressed that the robot should not ``\textit{run into the resident}'' and that it ``\textit{will try to avoid things.}''

\paragraph{Sensing abilities.}
Caregivers desired robots with multiple sensory abilities, soft sound detection, and situational awareness. While the caregivers mentioned standard capabilities such as vision, speech, hearing, and mobility, they also mentioned less common sensing abilities, including smell to monitor the environment, taste to help with cooking, and touch to discern if a material is wet. The envisioned uses for each ability are illustrated below:
\begin{quote}
    \textbf{AL3}:
    [The robot has] some sort of smell in case there's a fire or smoke, or a toaster, [or] a phone charger shorting out or something.
\end{quote}
\begin{quote}
    \textbf{IL1}:
    Let's say the robot was cooking and ... she's told the robot the recipe ... and asking the robot, ``Well how does that taste? Too much salt, too much that?'' [the robot will] know.
\end{quote}
\begin{quote}
    \textbf{AL3}:
    With laundry, touch might be [important]. If they can sense the dampness of the clothes ... Sometimes you can't see the soiling but you can feel the dampness.
\end{quote}

Additionally, from the field notes, we saw that when the caregiver knocks on a resident's door, the caregiver had to listen carefully to discern quiet responses from residents.

Lastly, both the field notes and interviews indicate a need for the robot to have social awareness. The field notes report that, at times, residents would stare at the caregivers to get their attention, described as ``prior-to-request behaviors'' in prior work \cite{yamazaki2007prior}. They may not actively seek attention but more passively waiting for the caregiver to come to them. 

\subsubsection{Theme 5: Providing Mental and Emotional Support}
The caregivers viewed their job as more than just the physical assistance they provide to residents in daily activities, emphasizing their role in providing mental and emotional support that residents ``\textit{need}'' (IL3). In light of the importance of this social support, the caregivers questioned whether a robot should provide such social support.

\paragraph{Companionship. }
Companionship is a significant part of the caregivers' interactions with residents, and it is evident that they formed close bonds as a result of their regular interactions. 
IL2 notes that the residents ``\textit{look forward to seeing someone}'' and that working with the residents over time is ``\textit{like being part of the family.}'' 
Because of this bond, residents will ``\textit{open up}'' (IL1) about their personal life. IL1 describes her relationship with the residents, saying ``\textit{I know a lot about them, and even sometimes when they're having problems or issues, they'll talk to me about it. And I'm just there to listen to them ... and if I can help in any way, I will.}''
This bond was observed throughout all observation sessions. The caregivers were constantly engaging in small talk and personal discussions with the residents, such as a resident commenting on IL1's change of hairstyle, or a resident revealing their personal goals to B1.

\paragraph{Comfort. }
In addition to companionship, comforting the resident appeared central to the caregivers' interactions with residents. IL2 describes an instance where a resident was distraught over some maintenance issues in her apartment and needed comfort:
\begin{quote}
    \textbf{IL2}:
    The resident that I gave the shower to today was talking about being nervous. And I was like, ``Just relax, it's okay. Take a deep breath. That's what we're here for. And if you ever need help early in the morning, I get here at 7. Call [with] your help button and I'll come in right away.'' So trying to make [the residents] as comfortable as possible, because this is their home. This is where they live, and so we want them to be happy.
\end{quote}
Another way the caregivers connect with and comfort their residents is by incorporating physical touch in the day-to-day interactions with them. This observation comes from the field notes --- for example, IL1 was seen using physical touch to comfort a resident.

\paragraph{Awareness. }
Specifically regarding a care robot, IL3 mentioned that the robot should be able to infer the resident's mental state, such as a resident's ``\textit{level of excitement}.'' This inference is important because it affects what the caregiver or robot will do next. IL3 envisions a scenario where the robot might need to adjust its plan upon seeing a distressed resident, saying ``\textit{[The robot] can read that [the resident] not calm so [the robot] cannot help her right now, [it] need[s] to help her to calm down first.}''

\paragraph{Concerns. }
While the caregivers were ``\textit{intrigued}'' (AL3) with the notion of having a robot to interact with residents, they expressed concerns about whether the robot would have that ``\textit{human factor}'' (IL1) in interactions. IL1 could not imagine ``\textit{a robot to sit down and give a resident comfort}'' because it ``\textit{doesn't have a beating heart.}'' AL3 worried care robots could create a ``\textit{colder society,}'' and stated that a care robot would have a challenging time with ``\textit{the empathy, the compassion, or friendship.}'' Instead, the caregivers were more interested in a robot ``\textit{assistant}'' to help them have more time to address these social needs of the residents. AL3 explains this idea saying ``\textit{If [robots] can lighten my load a little bit, and I can do more things that matter ... that's a good thing.}''

\subsubsection{Theme 6: Expectations of Interaction Modality}
Caregivers had a range of ideas about how the robot would know what to do. 
An idea shared among many was for the robot to be ``\textit{voice activated}'' (IL1) such that caregivers could simply ``\textit{tell it what to do}'' (AL2) and it would ``\textit{be able to understand and reply to questions or any demands}'' (IL2). AL3 describes how she would prefer the interaction to flow, saying ``\textit{I like the idea of the robot even being able to say `Go check on Mrs. Jones in 307.' Tell them 20 minutes, that I'm busy.'}\thinspace''
In addition to voice commands from caregivers, ``\textit{the resident would tell [the robot] what to do}'' (IL1). 

\paragraph{Programming. }
Another other form of interaction caregivers discussed was the ability to program the robot. Caregivers expressed interest in asking the robot to perform certain tasks or fill-in for a human caregiver as needed. 
AL3 caregiver expresses her vision to program the robot to check on residents throughout the day:
\begin{quote}
    \textbf{AL3}:
    I think would be easy to program it to do certain rounds. ... At 1:30[PM], go check for laundry. At 2:00[PM], go check [if personal care] products [are running low] and if they're needing anything. At 3:00[PM], just do a simple eyeball checker, you know or auditory or visual check on the resident to make sure they're okay. At 4:00[PM], set the table.
\end{quote}
However, AL3 later added to that vision, describing a hybrid approach where the pre-programmed robot would have to handle interruptions in emergent situations:
\begin{quote}
    \textbf{AL3}: 
    I think a lot could be programmed, but obviously I think ... the [caregiver] would be able to interject at some point. Because let's say there's six calls going and someone just fell. Then you're gonna be able to ... say ``Roll back, go check in room 307 ... They were supposed to go to bed 20 minutes ago. Are they safe?'' Or just notify [the resident that the caregiver is late.]
\end{quote}

\paragraph{Hierarchy. }
When discussing commanding the robot, however, all but one of the caregivers indicated that the robot should follow a hierarchy of authority. AL3 felt that a nurse manager ``\textit{would ... need to program it, [as far as] what do we need to prioritize}'' but that it should ``\textit{also be sensitive to the [caregivers'] needs as things come up.}''
IL3 mentioned that perhaps the robot should not do everything it might be asked to. She provides an extreme example, saying ``\textit{I know it's gonna be hard because ... we had a resident and she said `I want somebody to kill me' ... Imagine if you had this ... then [the robot is] not gonna do it.}''
AL3 and IL2 voiced that perhaps the robot would need some oversight from the caregivers when performing critical tasks that might fail or harm the residents. During mealtime, AL3 thought that ``\textit{a staff person would have to ... do a double check to make sure that there's no deviance}'' from the diet that each resident should follow, such as ``\textit{low salt}'' or ``\textit{thickened}'' liquids. IL2 felt that in the event that the robot has ``\textit{malfunctions or something went wrong,}'' the caregiver would ``\textit{stop}'' the robot and ``\textit{show [it] the correct way of doing certain things.}''

\section{Discussion}
Our study seeks to understand caregiving workflows and practices and caregiver expectations of assistive care robots. Many of our results highlight a need for highly capable robots to act as a ``coworker,'' which aligns with results presented by \citet{sauppe2015social}. While we noted numerous differences in caregiving practices between AL and IL settings, AL and IL caregivers did not envision an assistive care robot differently. This lack of difference may be due to the caregivers' lack of familiarity with the capabilities of an assistive robot or because the tasks where they need assistance from a robot across the two settings are similar. Nonetheless, the results provide valuable insights into care practices and reveal promising opportunities for future design of care robots.

\begin{table*}[!hb]
\caption{A summary of the design implications as guidelines for future care robot design.}
\label{tab:design_guidelines}
\centering\small\renewcommand{\arraystretch}{1.1}
\begin{tabular}{p{0.13\linewidth}p{0.41\linewidth}p{0.41\linewidth}} 
    \toprule
    \textbf{Implication} & \textbf{Guideline} & \textbf{Example} \\ 
    \midrule
    \textit{Support} \\ 
    \addlinespace[.15cm]
    Capabilities & Robots should have multiple capabilities such as physically supporting residents, manipulating items, and proactive monitoring. &  A robot could lift residents in and out of bed, but also monitor for falls or other assistance that the resident needs. \\
    Control hierarchy & Robots should report to caregiver directly and clear resident requests with caregivers prior to performing them. & If a resident asks the robot for candy, the robot should confirm with the caregiver whether it can give candy to the resident. \\
    \midrule
    \textit{Customization} \\
    \addlinespace[.15cm]
    Caregiver-specified & Caregivers need to be able to express their domain knowledge of resident needs and preferences to the robot. & The caregiver should be able to set wake-up times, meal times, and drink preferences for each resident.\\
    Learned & Robots should adapt over time from input from caregivers and interacting with residents. & If each time the robot tries to deliver water to a resident in the morning, the resident is still asleep, the robot should adjust the time it will deliver the water to after the resident wakes up. 
    \\
    \midrule
    \textit{Acceptability} \\
    \addlinespace[.15cm]
    Social Awareness & Robots must be socially aware of the environment to respond appropriately to the resident's current state. & If a robot tries to deliver a snack to a resident, but that resident is expressing confusion about the robot and feeling unsafe, the robot should not simply leave the snack, but instead alert the caregiver that the resident is in need of human assistance.\\
    Transparency & Robot actions should be understandable to the resident to maintain their autonomy and clear to the caregiver for easy coordination and supervision. & The robot could inform residents about the actions it is performing and maintain a log of the tasks completed, so that the caregiver can verify the status of the robot's scheduled tasks. \\
    \bottomrule
\end{tabular}
\end{table*}

While care robots hold great promise, we must consider how the introduction of these technologies will affect caregivers' burdens. Care robots will introduce new responsibilities such as assigning tasks to the robot, troubleshooting errors, maintaining the robot, and coordinating robot use among multiple caregivers and residents. These additional demands on caregivers must be offset by care robots that can effectively ease their \textit{objective} burden (i.e., care tasks the caregiver must perform).
Care robots might also alleviate caregivers' \textit{subjective} burden (i.e., the emotional toll that comes with providing the care). Previous work by \citet{wada2005psychological} found that interacting with a socially assistive robot long-term improved residents' moods, 
although similar long-term effects on caregivers are unclear. 
Considering both the objective and subjective burdens of caregivers will be critical to future design of care robots.

We also consider how our findings from professional caregivers relate to the needs identified by previous work in informal caregiving. 
Although they have different reasons, both formal and informal caregivers struggle with time. Formal caregivers balance multiple residents' needs, and informal caregivers balance caregiving with their personal lives \cite{chen2013caring}. The care robots envisioned in our work could also benefit informal caregivers by relieving some of their objective burdens. Robots such as Hobbit \cite{bajones2018hobbit} are already being developed for in-home fall monitoring, but adding manipulation capabilities would allow a care robot to complete simple tasks (e.g., fetching food or medicine and picking items up that were dropped) without needing the informal caregiver to be present. 

Current robots are not sufficiently capable for care settings, but their abailities are advancing. For example, Moxi\footnote{Moxi from Diligent: \url{https://www.diligentrobots.com/}} makes autonomous deliveries in hospitals. Recent work by \citet{odabasi2022refilling} shows that robots can perform simple tasks in care settings, while highlighting current limitations in perception, manipulation, and navigation.
Although robots such as Tiago \cite{pages2016tiago} can overcome some of these limitations, the ability of these bulky and expensive robots to find widespread adoption is unknown. 
Further, few robots are strong enough to lift humans. For example, the RIBA robot \cite{mukai2010development} is able to lift a person, but it has not yet been widely used in care settings. 
Although robots have a long way to go, advancing capabilities make the vision of care robots much more within reach. It is therefore critical to inform the development of these capabilities with an understanding of how care robots fit into the workflows of caregivers to more effectively focus development efforts and to facilitate future adoption.

\subsection{Design Implications}
We present a set of design implications that identify opportunities for care robots to support caregiver workflows and practices. Each implication is summarized as a guideline in Table \ref{tab:design_guidelines}.

\subsubsection{Support}
Caregivers envisioned care robots as \textit{assistants} that they could assign tasks to, enabling caregivers to engage residents in more meaningful ways. Care robots must therefore support the caregivers' existing workflows and needs that we describe in Theme 1 and in Theme 2, respectively. We combine these results with the physical capabilities of robots discussed in Theme 4 and the idea of a hierarchy from Theme 6 to identify two ways that robots can support caregivers: physical capabilities of the robot and its ability to fit into a hierarchical structure.

\paragraph{Capabilities.}
Caregivers indicated that care robots should serve as their \textit{assistants}, providing robust physical assistance and monitoring support to residents.
Incorporating multiple functions and abilities into a single robot raises important considerations for physical human-robot interaction.
Care machines today are only suitable for a specific task, such as lifting a person, cleaning a spill, giving a bath, or manipulating light items.
Despite recent advancements in areas such as \textit{soft robotics} \cite{whitesides2018soft}, creating strong, multipurpose robots that are also safe for elderly residents remains a challenge. 

\begin{figure}[!t]
    \centering
    \includegraphics[width=\columnwidth]{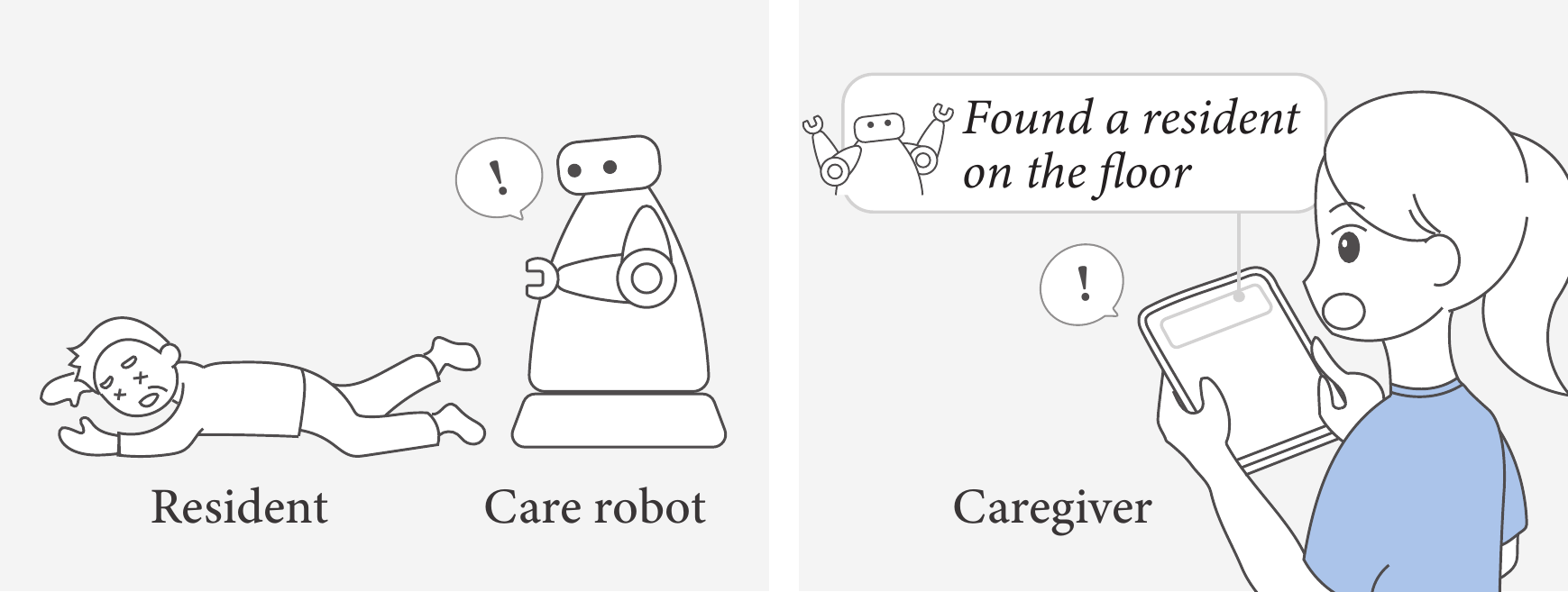}
    \caption{\textit{Left:} care robot identifies a fallen resident. \textit{Right:} care robot alerts the caregiver.
    }
    \Description{On the left, an elderly man is laying on the floor and a robot is next to him with an exclamation mark by its head. On the right, a caregiver receives a notification from a tablet interface that says "Found a resident on the floor."}
    \label{fig:monitoring_resident}
\end{figure}

Caregivers also expressed the importance of monitoring the resident's environment to proactively solve issues, such as detecting the dampness of a resident's chair in case they had incontinence. While not dire, these events can significantly affect the resident's quality of life. We need robots that can embody advanced sensing capabilities to create more holistic monitoring systems. One way to expand sensing capabilities is to incorporate robots into Ambient Assisted Living (AAL) practices \cite{aced2015supporting,zulas2012caregiver}. While previous work has explored how companion robots can be connected with smart sensors in private homes \cite{noury2005ailisa,badii2009companionable}, we must consider how robots can proactively monitor and respond to events in group living settings. 
For example, care robots can be used to alleviate privacy concerns that arise with constant monitoring, since a robot can check on residents periodically while otherwise not having access to the space. Figure \ref{fig:monitoring_resident} shows a situation where the robot is checking on a resident and finds that they have fallen, so the robot signals the caregiver to address the emergency.

\begin{figure}[!b]
    \centering
    \includegraphics[width=\columnwidth]{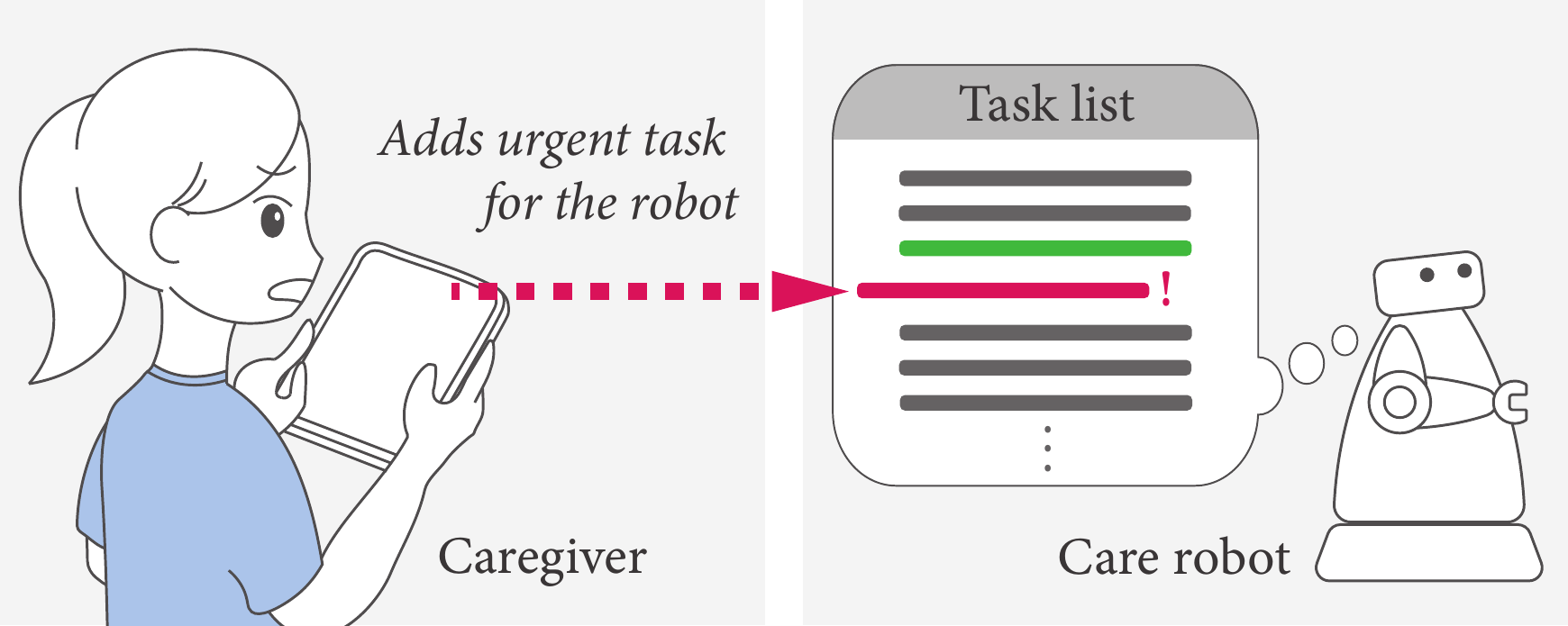}
    \caption{
   \textit{Left:} caregiver assigns an urgent task for the care robot to complete immediately. \textit{Right:} care robot interrupts its scheduled tasks to prioritize the caregiver's new task.
    }
    \Description{On the left, a caregiver holds a tablet. On the right, a robot stands with a list of its scheduled tasks. An arrow points from the caregiver's tablet to a new task in the robot's schedule.}
    \label{fig:caregiver_interruptions}
\end{figure}

\paragraph{Control hierarchy.}
Caregivers want to directly command the robot, and a few specifically mentioned concerns about to whom the robot will report and to what extend the robot should take input from residents.
In the case where multiple caregivers are working at the same time, the robot needs to manage multiple directives. Does the robot ``belong'' to a caregiver, such that it only listens to that caregiver unless temporarily handed off to another? We imagine a case where the robot is given a scheduled routine by a ``super user'' \cite{ernst2021human} but can accept on-the-fly input from other caregivers during a shift. Depending on what the robot is currently doing (idle, checking on residents, etc.) and what the caregiver has asked, the robot may adjust its schedule. An example scenario is shown in Figure \ref{fig:caregiver_interruptions}, where a caregiver asks the robot to interrupt its schedule to handle an urgent task. 
Care robots must handle task prioritization so that they can handle input from multiple sources.

The robot should also listen to input from the residents, but the goals and priorities of the caregiver and of the resident might conflict \cite{hasselkus1991ethical}.
Therefore, care robots need to manage potentially conflicting goals. 
Depending on the resident, overriding the caregiver's task may not be safe, such as the case of a resident who is hesitant to take medication or a diabetic resident who wants the robot to bring candy. These situations represent realistic ethical dilemmas that must be addressed. 
Recent research in this area includes models and proposals for integrating ethical principles into robot design \cite{sorell2014robot,malle2016integrating,stowers2016life,vanderelst2020can}.
One alternative defers to the caregiver --- the robot will query the caregiver if the resident asks the robot to perform actions that do not fit within the prescribed care. 
The robot should be designed to follow the care practices developed for the resident while respecting the resident's desire for autonomy by balancing control hierarchy and transparency. For example, if a diabetic patient asks for a piece of candy for morning snack, the robot could communicate to the resident that it has to run this request by the caregiver. The robot could also engage the caregiver in resolving the conflict between the request and the prescribed care and/or ask the caregiver for guidance on how to handle future requests by the resident. 
This approach maintains the robot's supporting role rather than allowing it to make decisions that could compromise resident care.
Since the robot is also learning what it can do for each resident over time, the robot will slowly refine its decision-making and reduce the workload of the caregiver.  However, not everything that the robot will learn will be the same. Safety-related tasks might be inflexible, whereas preference-based tasks should adapt over time.
As care robots become more capable, designers need to address the complicated dynamic that can emerge between conflicting caregiver and resident goals.

\subsubsection{Customization}
A common thread summarized in Theme 2 was that each resident has individual abilities, routines, and preferences, which supports previous findings on personalizing care robots \cite{beer2012domesticated,winkle2018social,law2019developing}. 
For example, a resident who has trouble hearing may require the robot to be closer and louder compared to a resident who is timid and prefers the robot to be at a distance. 

\begin{figure}[!b]
    \centering
    \includegraphics[width=\columnwidth]{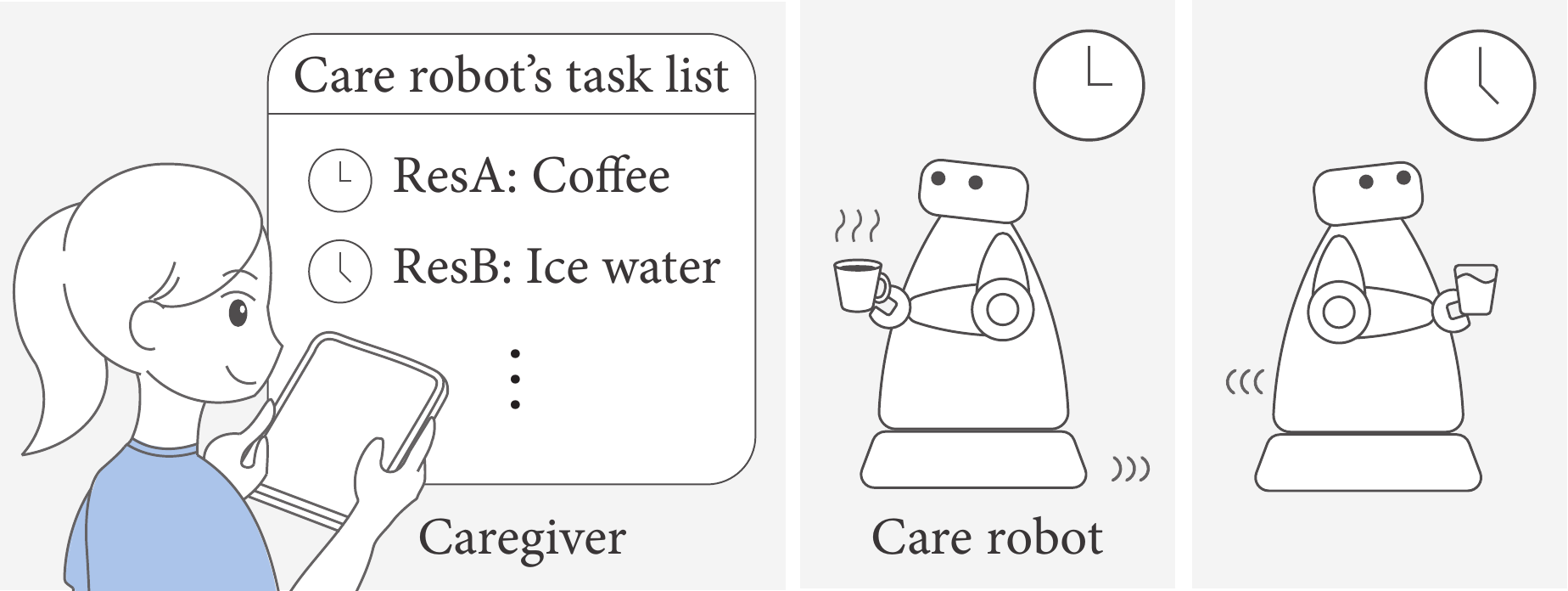}
    \caption{\textit{Left:} caregiver customizes drink delivery details for two residents. \textit{Center/Right:} care robot makes the deliveries. 
    }
    \Description{On the left, a caregiver holds a tablet. A thought bubble from the caregiver shows a list of two tasks: bring Res A coffee at x time, and bring Res B ice water at y time. On the right, the panel is split. First shows a robot at x time holding a cup of coffee, and second shows a robot at y time holding a cup of ice water.}
    \label{fig:caregiver_customization}
\end{figure}

\paragraph{Caregiver-specified customization.}
We need care robots that can be easily customized by caregivers, such as through end-user programming. Caregivers have extensive knowledge of the individual needs of residents, making them appropriate domain experts for customizing these care robots. The caregiver should be able to customize different robot behaviors for each resident and set a schedule of tasks for the robot to do, such as the scenario presented in Figure \ref{fig:caregiver_customization}.  
This recommendation is supported by results from a study by \citet{schroeter2013realization}, where caregivers in home settings expressed the desire to set up and control the robot. Recent applications of trigger-action programming for robots  \cite{leonardi2019trigger,senft2021situated} can be a fruitful avenue of exploration for end-user programming of care robots. Existing autonomous care robots, such as Hobbit \cite{fischinger2016hobbit}, use simple command interfaces that are suitable for basic interactions. Further interfaces and programming paradigms must be explored to enable care robots to follow more complex sequences of actions.

\paragraph{Learned customization.}
Care robots should also adapt to the needs and preferences of individual residents based on past interactions, such as through a combining learning techniques and formal verification. 
Reinforcement learning has shown promise for adapting robot social behaviors over time, particularly within the education \cite{gordon2016affective,park2019model} and service \cite{tseng2018active,chen2018information} domains.
\citet{porfirio2020transforming} additionally used formal verification to ensure that adapted programs adhere to social guidelines.
Verification techniques such as model checking have also been employed in the care setting to increase the trustworthiness of autonomous service robots \cite{webster2016toward,dixon2014fridge,webster2014formal}. Care robots should learn on their own while ensuring correct and safe behaviors to ease the burden on caregivers who would otherwise have to customize them.

\begin{figure}[!t]
    \centering
    \includegraphics[width=\columnwidth]{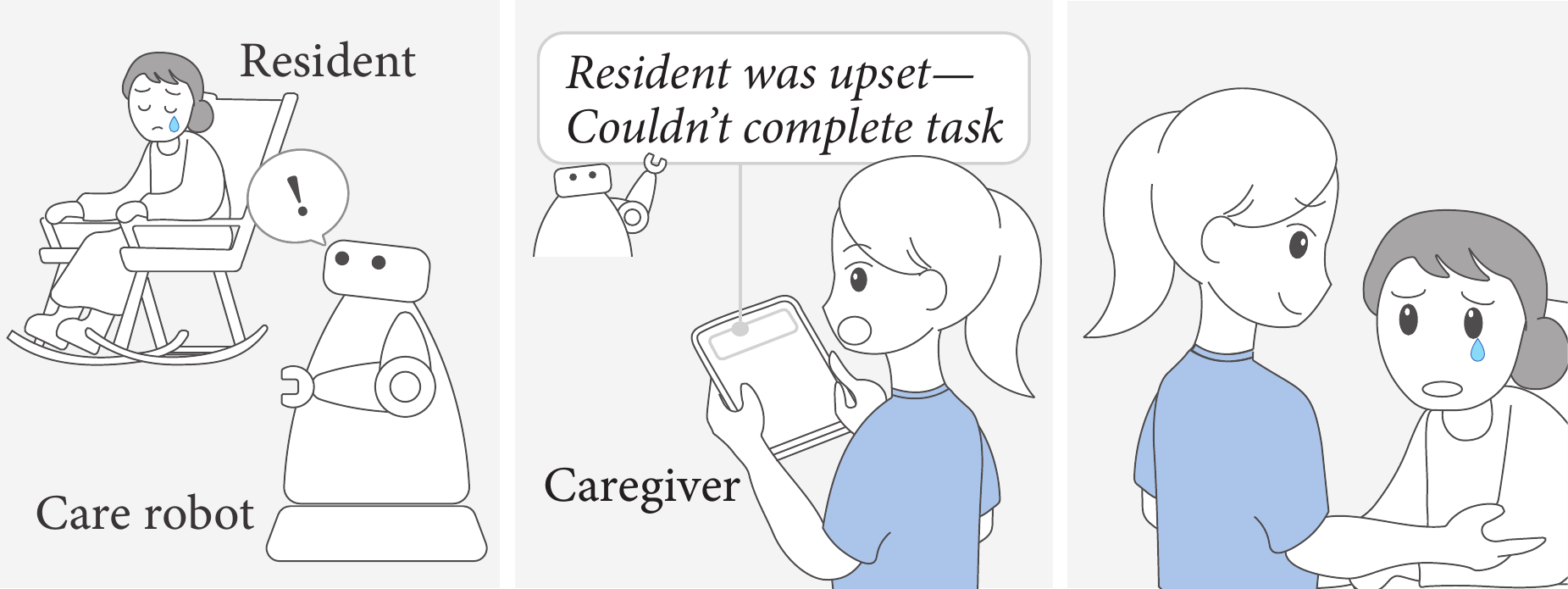}
    \caption{\textit{Left:} care robot arrives to help resident, but observes that the resident is upset. \textit{Center:} caregiver receives notification indicating that the upset resident needs assistance that the robot cannot provide. \textit{Right:} caregiver comforts the resident while the robot completes other tasks.
    }
    \Description{On the left, an elderly woman sits in a chair, crying. A robot is looking at her. In the center, a caregiver is holding a tablet with a notification that says "Resident was upset. Couldn't complete task." On the right, elderly woman is sitting in a chair with a tear in her eye. The caregiver is smiling and touching her hand.}
    \label{fig:appropriate_behavior}
\end{figure}

\subsubsection{Acceptability}
For care robots to be accepted in senior living communities, they must meet the expectations of residents and caregivers. We combine the ideas of social support from Theme 5 and the communication from Theme 3 to consider how robots can be acceptable through social awareness and promoting transparency.

\paragraph{Social Awareness.}
Caregivers emphasized the importance of social awareness because it allows them to respond appropriately to a resident's state. Upset or confused residents should be addressed differently than jovial or excited residents. 
Robots need to likewise respond appropriately to various resident states they encounter.
As care robots are viewed as social agents to residents in senior living communities \cite{gross2015robot}, introducing them to senior living communities creates a triadic interaction between the robot, caregiver, and resident. Whereas the dyadic model between resident and caregiver is clear (i.e., the resident has a need that the caregiver attends), the triadic model involving a robot is not well-developed. 
We recommend that the robot provides physical assistance as prescribed by the caregiver, but that it is also socially aware of the resident's state so that it can prompt the caregiver to assist when necessary. 
One example is shown in Figure \ref{fig:appropriate_behavior}, where the robot arrives to help the resident with a task, but the resident is upset. As a result, the robot is unable to complete the task, so it alerts the caregiver to assist the upset resident.
Although developing emotionally intelligent robots \cite{yan2021emotion} and socially assistive robots \cite{law2019developing} make up a significant body of research, we must find an acceptable balance of social assistance capabilities in physically assistive care robots.

\begin{figure}[!t]
    \centering
    \includegraphics[width=\columnwidth]{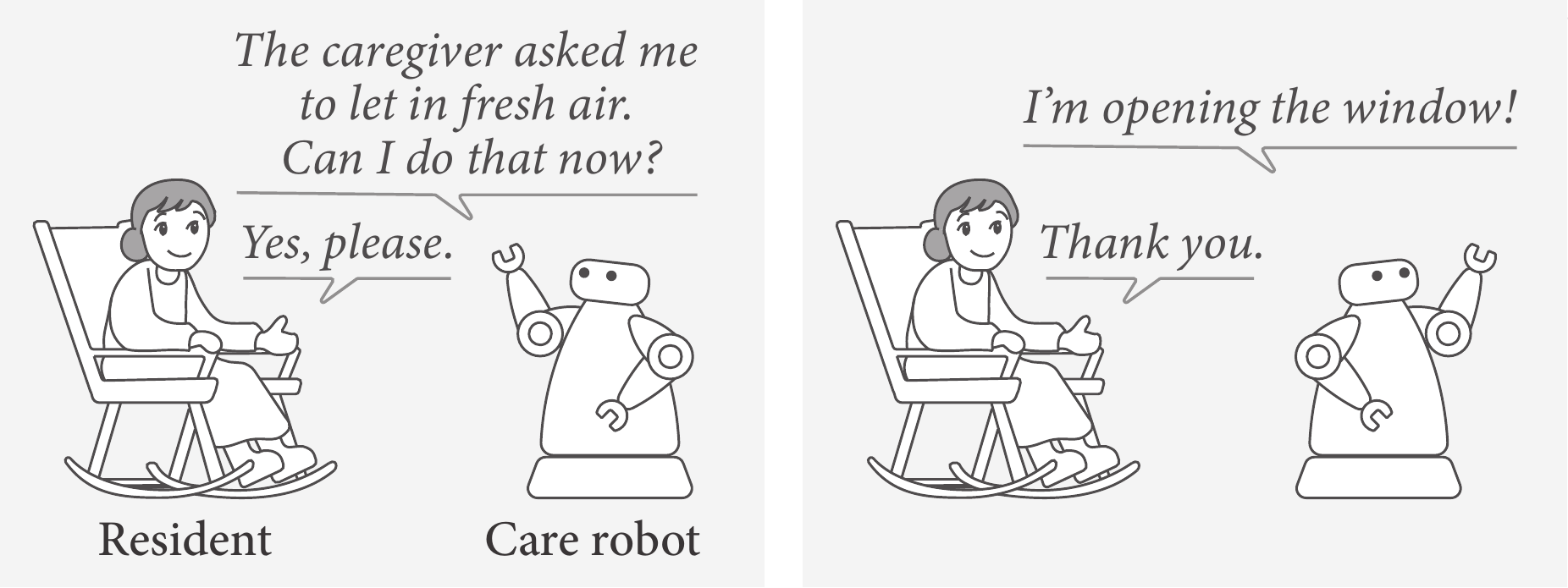}
    \caption{\textit{Left:} robot arrives to help resident, and asks permission to complete the task assigned by the caregiver. \textit{Right:} robot narrates its actions to keep the resident informed.
    }
    \Description{On the left, an elderly woman sits in a chair. A robot is next to her, saying "The caregiver asked me to let in fresh air. Can I do that now?" And the elderly woman says "Yes, please." On the right, an elderly woman sits in a chair. A robot is next to her, saying "I'm opening the window!" And the elderly woman says "Thank you."}
    \label{fig:promote_transparency}
\end{figure}

\paragraph{Transparency.}
Caregivers also maintain transparency with their actions and intentions when caring for residents, such as asking for consent before performing tasks or informing them of what is going on. Caregivers do so even if the task is straightforward or required by the caregiver (i.e., the resident cannot opt out). This interaction helps maintain resident autonomy by keeping them involved in their care. 
Care robots must continue to promote this transparency by embodying caregivers' transparency principles.
An example scenario is shown in Figure \ref{fig:promote_transparency}, where the robot is sent by the caregiver to open the window and clearly communicates its intentions to the resident.

Care robots must also be transparent to the caregivers about their actions. While the robot will not deviate heavily from its instructions, it is possible for the robot to take input from multiple caregivers or residents, or to use learning techniques to automatically refine its task performance. Therefore, the robot should maintain a human-readable care log, where it tracks each task and learned adaptations. 
To enable caregivers to stay up-to-date with a robot's autonomy and check that a robot is not learning undesirable behaviors, we can draw from \textit{explainable artificial intelligence} (XAI) \cite{gunning2019xai} to promote transparency and trustworthiness in the robot's automated learning approaches.

\subsection{Limitations and Future Work}
While our findings offer insights into how caregivers might use care robots in AL and IL settings, our study has two key limitations that must be addressed in future work. First, we only consider a small number of caregivers from one care facility and do not include perspectives from residents, their families, or non-caregiving facility staff or from stakeholders at other care facilities. 
Future work should expand on this preliminary study and seek to include more stakeholders and to understand how caregiver practices, workflows, and expectations vary across care facilities. 
While the skew in participant population (all female) was expected since the majority of professional caregivers are women \cite{argentum2018senior}, it does not account for the minority perspective of men. Future work should consider how the needs and perspectives of other types of caregivers and minorities differ.
Second, caregivers reflected on usage opportunities for robots based on images shown by the researcher but indicated that they must see and use the robots in person to provide more concrete ideas. Future work will be required to involve co-design sessions where caregivers experience controlling, programming, and interacting with one or more care robots to better understand their capabilities and limitations. 
\section{Conclusion}
We used ethnographic and co-design methods to explore design opportunities to support caregiving in senior living communities with robotic assistants. Our findings help characterize how caregivers for individuals with disabilities and age-related challenges work and understand how caregivers imagine a care robot could assist with their work. We provide design implications
organized in three different parts: supporting caregiver workflows, adapting to resident abilities, and providing feedback to all stakeholders of the interaction. To support caregiver workflows, care robots must have multiple capabilities such as physically supporting residents, manipulating items, and proactive monitoring. They must also fit into a control hierarchy where the robot relies on the caregiver to address conflicts between caregiver and resident desires.
To adapt to resident capabilities, caregivers need to be able to express their domain knowledge of resident needs and preferences to the robot, and robots should also adapt over time from input from caregivers and interactions with the residents.
To provide feedback to all stakeholders of the interaction, robots need to provide feedback to both the resident and caregiver so that all parties involved in the triadic interaction are aware of the robot status and intentions.
Our findings contribute to the growing body of work surrounding care robots, specifically by considering design opportunities from a caregiver perspective in senior living communities.

\begin{acks}
We would like to thank the caregivers and the facility for participating in our study, Emmanuel Senft for providing guidance and support throughout the writing process, Amy Koike for making the figures, and Hui-Ru Ho for conducting our reliability coding.

This material is based upon work supported by National Science Foundation (NSF) award IIS-1925043 and the NSF Graduate Research Fellowship Program under Grant No. DGE-1747503. 
Any opinions, findings, and conclusions or recommendations expressed in this material are those of the author(s) and do not necessarily reflect the views of the NSF.
\end{acks}

\bibliographystyle{ACM-Reference-Format}
\bibliography{ref}

\end{document}